\documentclass[10pt,twocolumn,letterpaper]{article}

\usepackage{cvpr}
\usepackage{times}
\usepackage{epsfig}
\usepackage{graphicx}
\usepackage{amsmath}
\usepackage{amssymb}

\usepackage[utf8]{inputenc} 
\usepackage[T1]{fontenc}    
\usepackage{url}            
\usepackage{booktabs}       
\usepackage{amsfonts}       
\usepackage{nicefrac}       
\usepackage{microtype}      

\usepackage{graphicx}
\usepackage{subcaption}
\usepackage{amsmath}
\usepackage{xcolor}
\usepackage[linesnumbered,ruled,vlined,onelanguage]{algorithm2e}
\usepackage{caption}
\usepackage{subcaption}
\usepackage{comment}
\usepackage{wrapfig,lipsum,booktabs}
\usepackage{amssymb,amsthm}
\usepackage{verbatimbox}
\usepackage{multirow}

\newcommand{\sys}{{\sc BFT }} 
\newcommand{\real}{\rm I\!R}

\usepackage[breaklinks=true,bookmarks=false]{hyperref}

\cvprfinalcopy 

\def\cvprPaperID{7976} 

\ifcvprfinal\fi
\begin{document}
\title{Butterfly Transform: An Efficient FFT Based Neural Architecture Design}

\author{Keivan Alizadeh vahid,\ Anish Prabhu,\ Ali Farhadi,\  Mohammad Rastegari \\
  University of Washington \\
  {\tt keivan@cs.washington.edu}
}

\maketitle
\thispagestyle{empty}
\begin{abstract}


   
   In this paper, we show that extending the butterfly operations from the FFT algorithm to a general Butterfly Transform (BFT) can be beneficial in building an efficient block structure for CNN designs.
   Pointwise convolutions, which we refer to as channel fusions, are the main computational bottleneck in the state-of-the-art efficient CNNs (e.g. MobileNets \cite{howard2017mobilenets,sandler2018mobilenetv2,mobilenet_v3_dblp}).We introduce a set of criterion for channel fusion, and prove that BFT yields an asymptotically optimal FLOP count with respect to these criteria. By replacing pointwise convolutions with BFT, we reduce the computational complexity of these layers from $\mathcal{O}(n^2)$ to $\mathcal{O}(n\log n)$ with respect to the number of channels.  
 Our experimental evaluations show that our method results in significant accuracy gains across a wide range of network architectures, especially at low FLOP ranges. For example, BFT results in up to a $6.75\%$ absolute Top-1 improvement for MobileNetV1\cite{howard2017mobilenets}, $4.4 \%$ for ShuffleNet V2\cite{ma2018shufflenet} and $5.4\%$ for MobileNetV3\cite{mobilenet_v3_dblp} on ImageNet under a similar number of FLOPS. Notably, ShuffleNet-V2+BFT outperforms state-of-the-art architecture search methods MNasNet\cite{tan2018mnasnet}, FBNet \cite{wu2018fbnet} and MobilenetV3\cite{mobilenet_v3_dblp} in the low FLOP regime.
\end{abstract}

\section{Introduction}
Devising Convolutional Neural Networks (CNN) that can run efficiently on resource-constrained edge devices has become an important research area. There is a continued push to put increasingly more capabilities on-device for personal privacy, latency, and scale-ability of solutions. On these constrained devices, there is often extremely high demand for a limited amount of resources, including computation and memory, as well as power constraints to increase battery life. Along with this trend, there has also been greater ubiquity of custom chip-sets, Field Programmable Gate Arrays (FPGAs), and low-end processors that can be used to run CNNs, rather than traditional GPUs.

A common design choice is to reduce the FLOPs and parameters of a network by factorizing convolutional layers~\cite{howard2017mobilenets, sandler2018mobilenetv2, ma2018shufflenet,zhang2017shufflenet} into a depth-wise separable convolution that consists of two components: (1) \emph{spatial fusion}, where each spatial channel is convolved independently by a depth-wise convolution, and (2) \emph{channel fusion}, where all the spatial channels are linearly combined by  $1 \times 1$ convolutions, known as pointwise convolutions. Inspecting the computational profile of these networks at inference time reveals that the computational burden of the spatial fusion is relatively negligible compared to that of the channel fusion\cite{howard2017mobilenets}. In this paper we focus on designing an efficient replacement for these pointwise convolutions.

\begin{figure}[t!]
\centering
\includegraphics[width=1.0\linewidth]{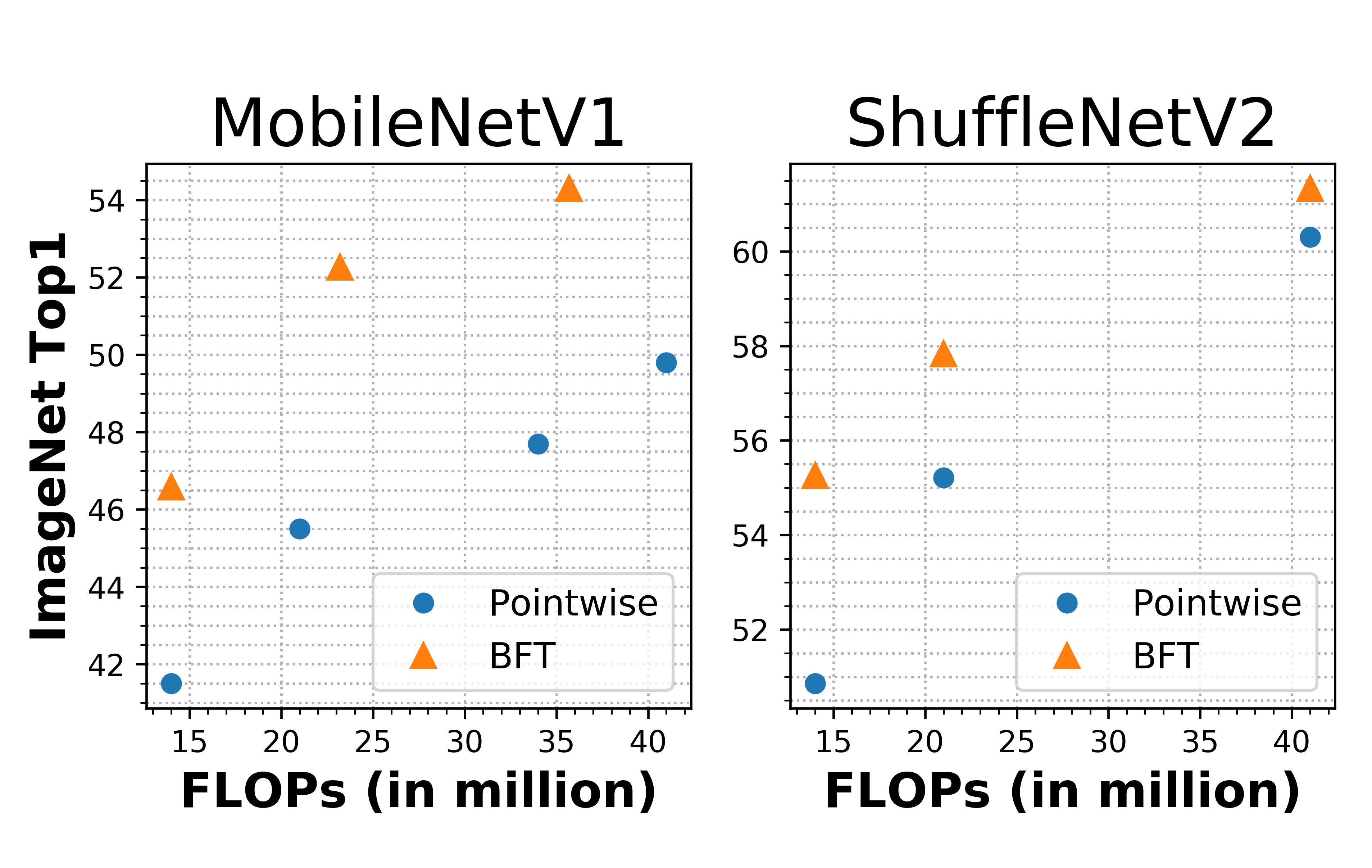}
\caption{\footnotesize Replacing pointwise convolutions with BFT in state-of-the-art architectures results in significant accuracy gains in resource constrained settings.}
\label{fig:mobilenet_spectrum}
\end{figure}

We propose a set of principles to design a replacement for pointwise convolutions motivated by both efficiency and accuracy. The proposed principles are as follows: (1) \textit{full connectivity from every input to all outputs}: to allow outputs to use all available information, (2) \textit{large information bottleneck}: to increase representational power throughout the network, (3) \textit{low operation count}: to reduce the computational cost, (4) \textit{operation symmetry}: to allow operations to be stacked into dense matrix multiplications. In Section~\ref{sec:model}, we formally define these principles, and mathematically prove a lower-bound of $O(n\log n)$ operations to satisfy these principles. We propose a novel, lightweight convolutional building block based on the Butterfly Transform (BFT). We prove that BFT yields an asymptotically optimal FLOP count under these principles.

We show that BFT can be used as a drop-in replacement for pointwise convolutions in several state-of-the-art efficient CNNs. This significantly reduces the computational bottleneck for these networks. For example, replacing pointwise convolutions with BFT decreases the computational bottleneck of MobileNetV1 from $95\%$ to $60\%$, as shown in Figure \ref{fig:mobilenet_pie}. We empirically demonstrate that using BFT leads to significant increases in accuracy in constrained settings, including up to a $6.75\%$ absolute Top-1 gain for MobileNetV1, $4.4\%$ for ShuffleNet V2 and $5.4\%$ for MobileNetV3 on the ImageNet\cite{deng2009imagenet} dataset. There have been several efforts on using butterfly operations in neural networks \cite{eunn_rnn,dao2019learning,Munkhoeva_quadrature} but, to the best of our knowledge, our method outperforms all other structured matrix methods (Table \ref{tab:lowrank}) for replacing pointwise convolutions as well as state-of-the-art Neural Architecture Search (Table \ref{tab:search}) by a large margin at low FLOP ranges.

\section{Related Work}
\label{sec:related_work}
Deep neural networks suffer from intensive computations. Several approaches have been proposed to address efficient training and inference in deep neural networks.

\paragraph{Efficient CNN Architecture Designs:} Recent successes in visual recognition tasks, including object classification, detection, and segmentation, can be attributed to exploration of different CNN designs \cite{lecun1990handwritten,simonyan2014very,he2016deep, krizhevsky2012imagenet,szegedy2015going, huang2017densely}. To make these network designs more efficient, some methods have factorized convolutions into different steps, enforcing distinct focuses on spatial and channel fusion \cite{howard2017mobilenets, sandler2018mobilenetv2}. Further, other approaches extended the factorization schema with sparse structure either in channel fusion \cite{ma2018shufflenet,zhang2017shufflenet} or spatial fusion \cite{mehta2018espnetv2}. \cite{huang2017condensenet} forced more connections between the layers of the network but reduced the computation by designing smaller layers. Our method follows the same direction of designing a sparse structure on channel fusion that enables lower computation with a minimal loss in accuracy. 

\paragraph{Structured Matrices:} There have been many methods which attempt to reduce the computation in CNNs, \cite{wen2016learning, li2018constrained, denton2014exploiting, jaderberg2014speeding} by exploiting the fact that CNNs are often extremely overparameterized. These models learn a CNN or fully connected layer by enforcing a linear transformation structure during the training process which has less parameters and computation than the original linear transform. Different kinds of structured matrices have been studied for compressing deep neural networks, including circulant matrices\cite{ding2017c}, toeplitz-like matrices\cite{toeplitz_small_footprint}, low rank matrices\cite{Sainath2013LowrankMF}, and fourier-related matrices\cite{ACDC_moczulski}. These structured matrices have been used for approximating kernels or replacing fully connected layers.
UGConv \cite{ugconv} has considered replacing one of the pointwise convolutions in the ShuffleNet structure with unitary group convolutions, while our Butterfly Transform is able to replace all of the pointwise convolutions.
The butterfly structure has been studied for a long time in linear algebra \cite{Parker95randombutterfly, Li2015ButterflyF} and neural network models \cite{bidirectional_bft}. Recently, it has received more attention from researchers who have used it in RNNs \cite{eunn_rnn}, kernel approximation\cite{Munkhoeva_quadrature, mathew_approx_hessian, choromanski-orthogmontcarlo} and fully connected layers\cite{dao2019learning}. 
We have generalized butterfly structures to replace pointwise convolutions, and have significantly outperformed all known structured matrix methods for this task, as shown in Table \ref{tab:lowrank}.


\paragraph{Network pruning:} This line of work focuses on reducing the substantial redundant parameters in CNNs by pruning out either neurons or weights \cite{han2015deep, han2015learning, Wortsman_neurips19, BagherinezhadRF16}. Our method is different from these type methods in the way that we enforce a predefined sparse channel structure to begin with and we do not change the structure of the network during the training.

\paragraph{Quantization:} Another approach to improve the efficiency of the deep networks is low-bit representation of network weights and neurons using quantization  \cite{soudry2014expectation,rastegari2016xnor,wu2016quantized,courbariaux2016binarized,zhou2016dorefa,hubara2016quantized,andri2018yodann}. These approaches use fewer bits (instead of 32-bit high-precision floating points) to represent weights and neurons for the standard training procedure of a network. In the case of extremely low bitwidth (1-bit) \cite{rastegari2016xnor} had to modify the training procedure to find the discrete binary values for the weights and the neurons in the network. Our method is orthogonal to this line of work and these method are complementary to our network. 

\begin{figure*}[t!]
    \centering
    \includegraphics[width=1.0\textwidth]{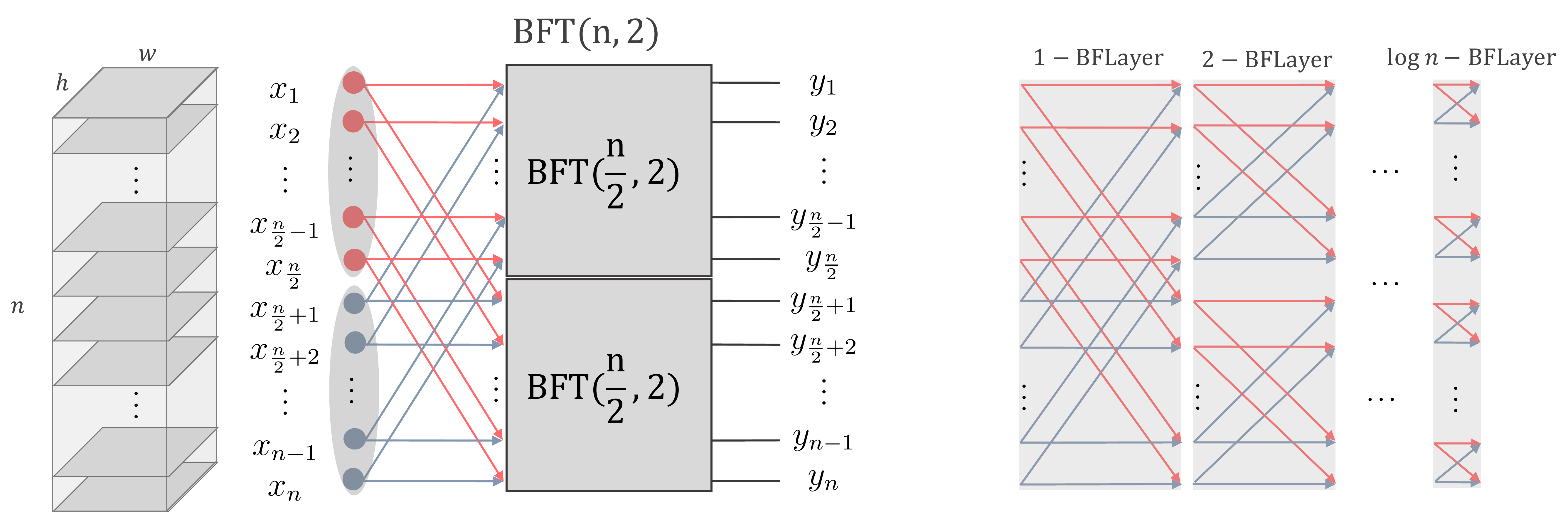}
    \caption{\footnotesize \textbf{BFT Architecture:} This figure illustrates the graph structure of the proposed Butterfly Transform. The left figure shows the recursive procedure of the BFT that is applied to an input tensor and the right figure shows the expanded version of the recursive procedure as $\log n$ Butterfly Layers in the network. }
    \label{fig:BFT}
\end{figure*}

\paragraph{Neural architecture search:} Recently, neural search methods, including reinforcement learning and genetic algorithms, have been proposed to automatically construct network architectures \cite{zoph2016neural,xie2017genetic,real2017large,zoph2018learning,tan2018mnasnet,liu2018progressive}. Recent search-based methods \cite{tan2018mnasnet,cai2018proxylessnas,wu2018fbnet, mobilenet_v3_dblp} use Inverted Residual Blocks \cite{sandler2018mobilenetv2} as a basic search block for automatic network design. The main computational bottleneck in most of the search based method is in the channel fusion and our butterfly structure does not exist in any of the predefined blocks of these methods. Our efficient channel fusion can be augmented with these models to further improve the efficiency of these networks. Our experiments shows that our proposed butterfly structure outperforms recent architecture search based models on small network design.

\section{Model}
\label{sec:model}

In this section, we outline the details of the proposed model. As discussed above, the main computational bottleneck in current efficient neural architecture design is in the channel fusion step, which is implemented with a pointwise convolution layer. The input to this layer is a tensor $\mathbf{X}$ of size $n_{\text{in}}\times h \times w$, where $n$ is the number of channels and $w$, $h$ are the width and height respectively.  The size of the weight tensor $\mathbf{W}$ is $ n_{\text{out}} \times n_\text{in} \times 1 \times 1 $ and the output tensor $\mathbf{Y}$ is $n_\text{out} \times h \times w$. For the sake of simplicity, we assume $n = n_\text{in} = n_\text{out}$. The complexity of a pointwise convolution layer is $\mathcal{O}(n^2wh)$, and this is mainly influenced by the number of channels $n$. We propose to use \emph{Butterfly Transform} as a layer, which has $\mathcal{O}((n\log n) wh)$ complexity. This design is inspired by the Fast Fourier Transform (FFT) algorithm, which has been widely used in computational engines for a variety of applications and there exist many optimized hardware/software designs for the key operations of this algorithm, which are applicable to our method. In the following subsections we explain the problem formulation and the structure of our butterfly transform.

\subsection{Pointwise Convolution as Matrix-Vector Products}
A pointwise convolution can be defined as a function $\mathcal{P}$ as follows:
\begin{equation}
    \mathbf{Y} = \mathcal{P}(\mathbf{X}; \mathbf{W})
\label{eq:pointwiseconv}    
\end{equation}
This can be written as a matrix product by reshaping the input tensor $\mathbf{X}$ to a 2-D matrix $\mathbf{\hat{X}}$ with size $n \times (hw)$ (each column vector in the $\mathbf{\hat{X}}$ corresponds to a spatial vector $\mathbf{X}[:,i,j]$) and reshaping the weight tensor to a 2-D matrix $\mathbf{\hat{W}}$ with size $n \times n$,  
\begin{equation}
    \mathbf{\hat{Y}} = \mathbf{\hat{W}}\mathbf{\hat{X}}
\end{equation}
where $\mathbf{\hat{Y}}$ is the matrix representation of the output tensor $\mathbf{Y}$. This can be seen as a linear transformation of the vectors in the columns of $\mathbf{\hat{X}}$ using $\mathbf{\hat{W}}$ as a transformation matrix. The linear transformation is a matrix-vector product and its complexity is $\mathcal{O}(n^2)$. By enforcing structure on this transformation matrix, one can reduce the complexity of the transformation. However, to be effective as a channel fusion transform, it is critical that this transformation respects the desirable characteristics detailed below.
 
\paragraph{Fusion network design principles:}
1) \textit{full connectivity from every input to all outputs}: This condition allows every single output to have access to all available information in the inputs. 2) \textit{large information bottleneck}: The bottleneck size is defined as the minimum number of nodes in the network that if removed, the information flow from input channels to output channels would be completely cut off (i.e. there would be no path from any input channel to any output channel). The representational power of the network is bound by the bottleneck size. To ensure that information is not lost while passed through the channel fusion, we set the minimum bottleneck size to $n$. 3) \textit{low operation count}: The fewer operations, or equivalently edges in the graph, that there are, the less computation the fusion will take. Therefore we want to reduce the number of edges. 4) \textit{operation symmetry}: By enforcing that there is an equal out-degree in each layer, the operations can be stacked into dense matrix multiplications, which is in practice much faster for inference than sparse computation.


\textbf{\textit{Claim}:} A multi-layer network with these properties has at least $\mathcal{O}(n \log n)$ edges. 

\textbf{\textit{Proof}:} Suppose there exist $n_i$ nodes in $i^\text{th}$ layer. Removing all the nodes in one layer will disconnect inputs from outputs. Since the maximum possible bottleneck size is $n$, therefore $n_i \geq n$.
Now suppose that out degree of each node at layer $i$ is $d_i$. Number of nodes in layer $i$, which are reachable from an input channel is $\prod_{j=0}^{i-1} d_j$. Because of the every-to-all connectivity, all of the $n$ nodes in the output layer are reachable. Therefore $\prod_{j=0}^{m-1} d_j \geq n$.
This implies that $\sum_{j=0}^{m-1} \log_2(d_j) \geq \log_2(n)$. The total number of edges will be:

$\sum_{j=0}^{m-1} n_j d_j \geq n\sum_{j=0}^{m-1} d_j \geq n \sum_{j=0}^{m-1} \log_2(d_j) \geq n\log_2n \blacksquare$

In the following section we present a network structure that satisfies all the design principles for fusion network. 

\subsection{Butterfly Transform (BFT)}
As mentioned above we can reduce the complexity of a matrix-vector product by enforcing structure on the matrix. There are several ways to enforce structure on the matrix. Here we first explain how the channel fusion is done through BFT and then show a family of the structured matrix equivalent to this fusion leads to a $\mathcal{O}(n\log n)$ complexity of operations and parameters while maintaining accuracy.  

\paragraph{Channel Fusion through BFT:}
We want to fuse information among all channels. We do it in sequential layers. In the first layer we partition channels to $k$ parts with size $\frac{n}{k}$ each, $\mathbf{x}_1,..,\mathbf{x}_k$. We also partition output channels of this first layer to $k$ parts with $\frac{n}{k}$ size each, $\mathbf{y}_1,..,\mathbf{y}_k$.
We connect elements of $\mathbf{x}_i$ to $\mathbf{y}_j$ with $\frac{n}{k}$ parallel edges $\mathbf{D}_{ij}$. After combining information this way, each $\mathbf{y}_i$ contains the information from all channels, then we recursively fuse information of each $\mathbf{y}_i$ in the next layers.
\paragraph{Butterfly Matrix:} 
In terms of matrices $\mathbf{B}^{(n,k)}$ is a butterfly matrix of order $n$ and base $k$ where $\mathbf{B}^{(n,k)}\in \real^{n\times n}$ is equivalent to fusion process described earlier.
\begin{equation}
    \mathbf{B}^{(n, k)} = 
    \begin{pmatrix}
    \mathbf{M}^{(\frac{n}{k},k)}_{1} \mathbf{D}_{11} & \dots & \mathbf{M}^{(\frac{n}{k},k)}_{1} \mathbf{D}_{1k}\\
    \vdots & \ddots & \vdots \\
    \mathbf{M}^{(\frac{n}{k},k)}_{k} \mathbf{D}_{k1} & \dots & \mathbf{M}^{(\frac{n}{k},k)}_{k} \mathbf{D}_{kk}\\
    \end{pmatrix}
\end{equation}

Where $\mathbf{M}^{(\frac{n}{k},k)}_{i}$ is a butterfly matrices of order $\frac{n}{k}$ and base $k$ and $\mathbf{D}_{ij}$ is an arbitrary diagonal $\frac{n}{k} \times \frac{n}{k}$ matrix. The matrix-vector product between a butterfly matrix $\mathbf{B}^{(n,k)}$ and a vector $\mathbf{x}\in \real ^{n}$ is :
\begin{equation}
    \mathbf{B}^{(n, k)}\mathbf{x} = 
    \begin{pmatrix}
    \mathbf{M}^{(\frac{n}{k},k)}_{1} \mathbf{D}_{11} & \dots & \mathbf{M}^{(\frac{n}{k},k)}_{1} \mathbf{D}_{1k}\\
    \vdots & \ddots & \vdots \\
    \mathbf{M}^{(\frac{n}{k},k)}_{k} \mathbf{D}_{k1} & \dots & \mathbf{M}^{(\frac{n}{k},k)}_{k} \mathbf{D}_{kk}\\
    \end{pmatrix}
    \begin{pmatrix}
    \mathbf{x}_1\\
    \vdots\\
    \mathbf{x}_k
    \end{pmatrix}
\end{equation}
where $\mathbf{x}_i \in \real ^{\frac{n}{k}}$ is a subsection of $\mathbf{x}$ that is achieved by breaking $\mathbf{x}$ into $k$ equal sized vector. Therefore, the product can be simplified by factoring out $\mathbf{M}$ as follow:
\begin{equation}
     \begin{matrix}
     \mathbf{B}^{(n, k)}\mathbf{x} = 
    \begin{pmatrix}
    \mathbf{M}^{(\frac{n}{k},k)}_1\sum_{j=1}^{k}{\mathbf{D}_{1j}\mathbf{x}_j}\\
    \vdots\\
    \mathbf{M}^{(\frac{n}{k},k)}_i\sum_{j=1}^{k}{\mathbf{D}_{ij}\mathbf{x}_j}\\
    \vdots\\
    \mathbf{M}^{(\frac{n}{k},k)}_k\sum_{j=1}^{k}{\mathbf{D}_{kj}\mathbf{x}_j}
    \end{pmatrix} = \begin{pmatrix}
    \mathbf{M}^{(\frac{n}{k},k)}_1\mathbf{y}_1\\
    \vdots\\
    \mathbf{M}^{(\frac{n}{k},k)}_i\mathbf{y}_i\\
    \vdots\\
    \mathbf{M}^{(\frac{n}{k},k)}_k\mathbf{y}_k
    \end{pmatrix}
    \end{matrix}
    \label{eq:halfbutterfly}
\end{equation}
where $\mathbf{y}_i = \sum_{j=1}^{k}{\mathbf{D}_{ij}\mathbf{x}_j}$. Note that $\mathbf{M}^{(\frac{n}{k},k)}_i\mathbf{y}_i$ is a smaller product between a butterfly matrix of order $\frac{n}{k}$ and a vector of size $\frac{n}{k}$ therefore, we can use divide-and-conquer to recursively calculate the product $\mathbf{B}^{(n, k)}\mathbf{x}$. If we consider $T(n,k)$ as the computational complexity of the product between a $(n,k)$ butterfly matrix and an $n$-D vector. From equation \ref{eq:halfbutterfly}, the product can be calculated by $k$ products of butterfly matrices of order $\frac{n}{k}$ which its complexity is $kT({n}/{k},k)$. The complexity of calculating $\mathbf{y}_i$ for all $i \in \{1, \dots ,k\}$ is $\mathcal{O}(kn)$ therefore:
\begin{equation}
       T(n,k) = kT({n}/{k},k)+\mathcal{O}(kn)\\
\end{equation}
\begin{equation}
    T(n,k) = \mathcal{O}(k(n\log_{k} n))
\end{equation}

With a smaller choice of $k (2\leq k \leq n)$ we can achieve a lower complexity. Algorithm \ref{alg:effproduct} illustrates the recursive procedure of a butterfly transform when $k=2$. 

\small
\newcommand{\forcond}{$i=0$ \KwTo $n$}
\SetKwFunction{BFT}{ButterflyTransform}%
\SetKwProg{Fn}{Function}{:}{}
\begin{algorithm}[!h]
\SetAlgoLined
\Fn(\tcc*[]{algorithm as a recursive function}){\BFT{W, X, n}}
{\KwData{W Weights containing $2n\log(n)$ numbers}
\KwData{X An input containing $n$ numbers}
\uIf{n == 1}{
    \KwRet{[X]} \;
  }
Make $D_{11}, D_{12}, D_{21}, D_{22}$ using first $2n$ numbers of $W$\;
Split rest $2n(\log(n)-1)$ numbers to two sequences $W_1, W_2$ with length $n(\log(n)-1)$ \;
Split $X$ to $X_1, X_2$\;
$y_1 \longleftarrow D_{11} X_1 + D_{12} X_2$\;
$y_2 \longleftarrow D_{21} X_1 + D_{22} X_2$\;

$My_1 \longleftarrow \BFT(W_1, y_1, n-1)$\;
$My_2 \longleftarrow \BFT(W_2, y_2, n-1)$\;
\KwRet{Concat($My_1, My_2$)}\;
}
\caption{Recursive Butterfly Transform}
\label{alg:effproduct}
\end{algorithm}


\textbf{\begin{figure*}[!t]
    \centering
    \includegraphics[width= 1.0\textwidth]{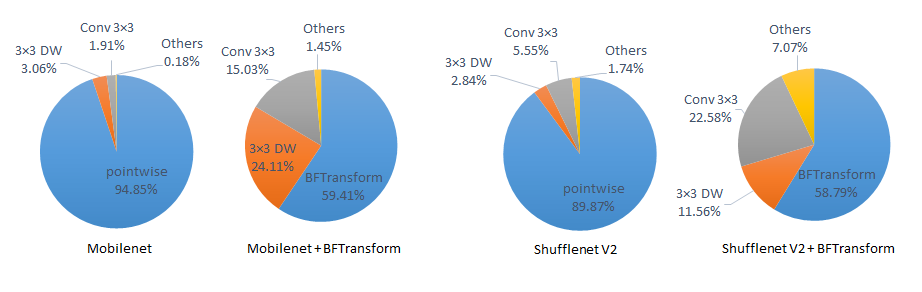}
    \caption {\footnotesize \textbf{Distribution of FLOPs:} This figure shows that replacing the pointwise convolution with BFT reduces the size of the computational bottleneck. }
    \label{fig:mobilenet_pie}
\end{figure*}}

\subsection{Butterfly Neural Network}

The procedure explained in Algorithm \ref{alg:effproduct} can be represented by a butterfly graph similar to the FFT's graph. The butterfly network structure has been used for function representation \cite{li2018butterfly} and fast factorization for approximating linear transformation \cite{dao2019learning}. We adopt this graph as an architecture design for the layers of a neural network. Figure ~\ref{fig:BFT} illustrates the architecture of a butterfly network of base $k=2$ applied on an input tensor of size $n\times h \times w$. The left figure shows how the recursive structure of the BFT as a network. The right figure shows the constructed multi-layer network which has $\log n$ Butterfly Layers (BFLayer). Note that the complexity of each Butterfly Layer is $\mathcal{O}(n)$ ($2n$ operations), therefore, the total complexity of the BFT architecture will be $\mathcal{O}(n\log n)$.

Each Butterfly layer can be augmented by batch norm and non-linearity functions (\textit{e.g.} ReLU, Sigmoid). In Section \ref{sec:ablation} we study the effect of using different choices of these functions. We found that both batch norm and nonlinear functions (ReLU and Sigmoid) are not effective within BFLayers. Batch norm is not effective mainly because its complexity is the same as the BFLayer $\mathcal{O}(n)$, therefore, it doubles the computation of the entire transform. We use batch norm only at the end of the transform. The non-linear activation $ReLU$ and $Sigmoid$ zero out almost half of the values in each BFLayer, thus multiplication of these values  throughout the forward propagation destroys all the information.         
The BFLayers can be internally connected with residual connections in different ways. In our experiments, we found that the best residual connections are the one that connect the input of the first BFLayer to the output of the last BFLayer. 
The base of the BFT affects the shape and the number of FLOPs. We have empirically found that base $k=4$ achieves the highest accuracy while having the same number FLOPs as the base $k=2$ as shown in Figure \ref{fig:butterfly_base}.

Butterfly network satisfies all the fusion network design principles. There exist exactly one path between every input channel to all the output channels, the degree of each node in the graph is exactly $k$, the bottleneck size is $n$, and the number of edges are $\mathcal{O}(n \log n)$. 
 
We use the BFT architecture as a replacement of the pointwise convolution layer ($1\times 1$ convs) in different CNN architectures including MobileNetV1\cite{howard2017mobilenets}, ShuffleNetV2\cite{ma2018shufflenet} and MobileNetV3\cite{mobilenet_v3_dblp}. Our experimental results shows that under the same number of FLOPs, the efficiency gain by BFT is more effective in terms of accuracy compared to the original model with smaller channel rate. We show consistent accuracy improvement across several architecture settings.    

Fusing channels using BFT, instead of pointwise convolution reduces the size of the computational bottleneck by a large-margin. Figure \ref{fig:mobilenet_pie} illustrate the percentage of the number of operations by each block type throughout a forward pass in the network. Note that when BFT is applied, the percentage of the depth-wise convolutions increases by $~8\times$.

\section{Experiments}
\label{sec:experiments}
In this section, we demonstrate the performance of the proposed \sys on large-scale image classification tasks. To showcase the strength of our method in designing very small networks, we compare performance of Butterfly Transform with pointwise convolutions in three state-of-the-art efficient architectures: (1) MobileNetV1, (2) ShuffleNetV2, and (3) MobileNetV3. We compare our results with other type of structured matrices that have $\mathcal{O}(n \log n)$ computation (e.g. low-rank transform and circulant transform). We also show that our method outperforms state-of-the art architecture search methods at low FLOP ranges.

\begin{table*}[!t]
\begin{small}
    \begin{subtable}{.50 \linewidth}
        \centering
        \caption{}
        \begin{tabular}{|l||l|l|l|}
            \hline
            Flops   & ShuffleNetV2 & ShuffleNetV2\textbf{+BFT} & Gain\\ \hline \hline
            14 M & 50.86 (14 M)* & 55.26 (14 M) & \textbf{4.40}\\ \hline
            21 M & 55.21 (21 M)* & 57.83 (21 M) & \textbf{2.62}\\ \hline
            40 M & \begin{tabular}[c]{@{}l@{}}59.70(41 M)*\\ 60.30 (41 M)\end{tabular} & 61.33 (41 M) & \begin{tabular}[c]{@{}l@{}}\textbf{1.63}\\\textbf{1.03}\end{tabular}\\ \hline
        \end{tabular}
        \label{tab:shufflenet_low_flop}
        \caption{}
        \begin{tabular}{|l||l|l|l|}
            \hline
            Flops   & MobileNetV3 & MobileNetV3\textbf{+BFT} & Gain \\ \hline \hline
            10-15 M & 49.8 (13 M) & 55.21 (15 M) & \textbf{5.41}\\ \hline
        \end{tabular}
    \label{tab:mobilenetv3_low_flop}

    \end{subtable}
    \begin{subtable}{.5\linewidth}
        \centering
        \caption{}
        \begin{tabular}{|l||l|l|l|}
            \hline
            Flops   & 
            MobileNet & MobileNet\textbf{+BFT} & Gain \\ \hline \hline
            14 M & 41.50 (14 M) & 46.58 (14 M) & \textbf{5.08}\\ \hline
            20 M & 45.50 (21 M) & 52.26 (23 M) & \textbf{6.76} \\ \hline
            40 M & \begin{tabular}[c]{@{}l@{}}47.70 (34 M) \\ 50.60 (41 M)\\ \end{tabular}& 54.30 (35 M) & \begin{tabular}[c]{@{}l@{}} \textbf{6.60} \\ \textbf{3.70} \end{tabular} \\ \hline
            50 M & 56.30 (49 M) & \begin{tabular}[c]{@{}l@{}}57.56 (51 M) \\ 58.35 (52 M)\\ \end{tabular} & \begin{tabular}[c]{@{}l@{}} \textbf{1.26} \\ \textbf{2.05} \end{tabular} \\ \hline
            110 M & 61.70 (110 M) & 63.03 (112 M)& \textbf{1.33} \\ \hline
            150 M & 63.30 (150 M) & 64.32 (150 M)& \textbf{1.02} \\ \hline
        \end{tabular}
                \label{tab:mobilenet_low_flop}

    \end{subtable}

\end{small}

    \label{tab:mobilenet_shufflenet}
    \caption{\footnotesize These tables compare the accuracy of ShuffleNetV2, MobileNetV1 and MobileNetV3 when using standard pointwise convolution vs using BFTs}
\end{table*}

\subsection{Image Classification}
\subsubsection{Implementation and Dataset Details:} 
Following standard practice, we evaluate the performance of Butterfly Transforms on the ImageNet dataset, at different levels of complexity, ranging from 14 MFLOPS to 150 MFLOPs. ImageNet classification dataset contains 1.2M training samples and 50K validation samples, uniformly distributed across 1000 classes.

For each architecture, we substitute pointwise convolutions with Butterfly Transforms. To keep the FLOP count similar between BFT and pointwise convolutions, we adjust the channel numbers in the base architectures (MobileNetV1, ShuffleNetV2, and MobileNetV3). For all architectures, we optimize our network by minimizing cross-entropy loss using SGD. Specific learning rate regimes are used for each architecture which can be found in the Appendix. Since BFT is sensitive to weight decay, we found that using little or no weight decay provides much better accuracy. We experimentally found (Figure \ref{fig:butterfly_base}) that butterfly base $k = 4$ performs the best. We also used a custom weight initialization for the internal weights of the Butterfly Transform which we outline below. More information and intuition on these hyper-parameters can be found in our ablation studies (Section \ref{sec:ablation}).


\paragraph{Weight initialization:}
Proper weight initialization is critical for convergence of neural networks, and if done improperly can lead to instability in training, and poor performance. This is especially true for Butterfly Transforms due to the amplifying effect of the multiplications within the layer, which can create extremely large or small values. A common technique for initializing pointwise convolutions is to initialize weights uniformly from the range $(-x, x)$ where $x = \sqrt{\frac{6}{n_{in}+n_{out}}}$, which is referred to as Xavier initialization~\cite{xavier-init}. We cannot simply apply this initialization to butterfly layers, since we are changing the internal structure.

We denote each entry $B^{(n,k)}_{u,v}$ as the multiplication of all the edges in path from node $u$ to $v$. We propose initializing the weights of the butterfly layers from a range $(-y,y)$, such that the multiplication of all edges along paths, or equivalently values in $B^{(n,k)}$, are initialized close to the range $(-x,x)$. To do this, we solve for a $y$ which makes the expectation of the absolute value of elements of $B^{(n,k)}$ equal to the expectation of the absolute value of the weights with standard Xavier initialization, which is $x/2$. Let $e_1,..,e_{log(n)}$ be edges on the path $p$ from input node $u$ to output node $v$. We have the following:
\begin{equation}
E[|B^{(n,k)}_{u,v}|] = E[|\prod_{i=1}^{log(n)}{e_i}|] = \frac{x}{2}
\end{equation}
We initialize each $e_i$ in range $(-y, y)$ where
\begin{equation}
(\frac{y}{2})^{log(n)} = \frac{x}{2} \implies y = x^{\frac{1}{log(n)}} * 2^{\frac{log(n)-1}{log(n)}}.
\end{equation}

\setlength{\intextsep}{5pt}%
\setlength{\columnsep}{7pt}%

\begin{table*}[!t]
\begin{small}

    \label{tab:othercomp}
    \begin{subtable}{.5\linewidth}
      \centering
        \caption{\textbf{BFT vs. Architecture Search}}
        \begin{tabular}{|l|l|}
\hline
Model               & Accuracy \\ \hline\hline
ShuffleNetV2+\textbf{BFT} (14 M) & \textbf{55.26}    \\ \hline
MobileNetV3Small-224-0.5+\textbf{BFT} (15 M) & \textbf{55.21}    \\ \hline
FBNet-96-0.35-1 (12.9 M) & 50.2    \\ \hline
FBNet-96-0.35-2 (13.7 M) & 51.9       \\ \hline
MNasNet (12.7 M) & 49.3     \\ \hline
MobileNetV3Small-224-0.35 (13 M) & 49.8     \\ \hline
MobileNetV3Small-128-1.0 (12 M) & 51.7     \\ \hline
\end{tabular}
        \label{tab:search}
    \end{subtable}%
    \begin{subtable}{.5\linewidth}
      \centering
        \caption{\textbf{BFT vs. Other Structured Matrix Approaches}}
        \begin{tabular}{|l|l|}
\hline
Model               & Accuracy \\ \hline\hline
MobilenetV1+\textbf{BFT} (35 M)       & \textbf{54.3}     \\ \hline
MobilenetV1 (42 M) & 50.6    \\ \hline
MobilenetV1+Circulant* (42 M) & 35.68    \\ \hline
MobilenetV1+low-rank* (37 M)  & 43.78       \\ \hline
MobilenetV1+BPBP (35 M)  & 49.65       \\ \hline
MobilenetV1+Toeplitz* (37 M)  & 40.09       \\ \hline
MobilenetV1+FastFood* (37 M)  & 39.22       \\ \hline
\end{tabular}
        \label{tab:lowrank}
    \end{subtable} 
    
    \end{small}
        \caption{\footnotesize These tables compare BFT with other efficient network design approaches. In Table (a), we show that ShuffleNetV2 + BFT outperforms state-of-the-art neural architecture search methods (MNasNet \cite{tan2018mnasnet}, FBNet\cite{wu2018fbnet}, MobilenetV3\cite{mobilenet_v3_dblp}). In Table (b), we show that BFT achieves significantly higher accuracy than other structured matrix approaches which can be used for channel fusion. The * denotes that this is our implementation.}

\end{table*}

\subsubsection{MobileNetV1 + BFT}

\begin{wrapfigure}{r}{3cm}
\centering
    \includegraphics[]{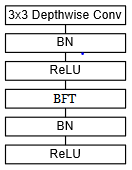}
    \caption {\footnotesize \\ \textbf{MobileNetV1+BFT Block}}
    \label{fig:mobilenet_bft_block}
\end{wrapfigure}

To add BFT to MobileNeV1, for all MobileNetV1 blocks, which consist of a depthwise layer followed by a pointwise layer, we replace the pointwise convolution with our Butterfly Transform, as shown in Figure \ref{fig:mobilenet_bft_block}. We would like to emphasize that this means we replace \emph{all} pointwise convolution in MobileNetV1, with BFT. In Table \ref{fig:mobilenet_spectrum}, we show that we outperform a spectrum of MobileNetV1s from about 14M to 150M FLOPs with a spectrum of MobileNetV1s+BFT within the same FLOP range. Our experiments with MobileNetV1+BFT include all combinations of width-multiplier 1.00 and 2.00, as well as input resolutions 128, 160, 192, and 224. We also add a width-multiplier 1.00 with input resolution 96 to cover the low FLOP range (14M). A full table of results can be found in the Appendix.


In Table \ref{tab:mobilenet_low_flop} we showcase that using BFT outperforms traditional MobileNets across the entire spectrum, but is especially effective in the low FLOP range. For example using BFT results in an increase of 6.75\% in top-1 accuracy at 23 MFLOPs. Note that MobileNetV1 + BFT at 23 MFLOPs has much higher accuracy than MobileNetV1 at 41 MFLOPs, which means it can get higher accuracy with almost half of the FLOPs. This was achieved without changing the architecture at all, other than simply replacing pointwise convolutions, which means there are likely further gains by designing architectures with BFT in mind.

\subsubsection{ShuffleNetV2 + BFT}
We modify the ShuffleNet block to add BFT to ShuffleNetv2. In Table \ref{tab:shufflenet_low_flop} we show results for ShuffleNetV2+BFT, versus the original ShuffleNetV2. We have interpolated the number of output channels to build ShuffleNetV2-1.25+BFT, to be comparable in FLOPs with a ShuffleNetV2-0.5. We have compared these two methods for different input resolutions (128, 160, 224) which results in FLOPs ranging from 14M to 41M. ShuffleNetV2-1.25+BFT achieves about 1.6\% better accuracy than our implementation of ShuffleNetV2-0.5 which uses pointwise convolutions. It achieves 1\% better accuracy than the reported numbers for ShuffleNetV2~\cite{ma2018shufflenet} at 41 MFLOPs.

\textbf{\subsubsection{MobileNetV3 + BFT}}

We follow a procedure which is very similar to that of MobileNetV1+BFT, and simply replace all pointwise convolutions with Butterfly Transforms. We trained a MobileNetV3+BFT Small with a network-width of 0.5 and an input resolution 224, which achieves $55.21\%$ Top-1 accuracy. This model outperforms MobileNetV3 Small network-width of 0.35 and input resolution 224 at a similar FLOP range by about $5.4\%$ Top-1, as shown in \ref{tab:mobilenetv3_low_flop}. Due to resource constraints, we only trained one variant of MobileNetV3+BFT.
\subsubsection{Comparison with Neural Architecture Search}
Including BFT in ShuffleNetV2 allows us to achieve higher accuracy than state-of-the-art architecture search methods, MNasNet\cite{tan2018mnasnet}, FBNet \cite{wu2018fbnet}, and MobileNetV3 \cite{mobilenet_v3_dblp} on an extremely low resource setting ($\sim$ 14M FLOPs). These architecture search methods search a space of predefined building blocks, where the most efficient block for channel fusion is the pointwise convolution. In Table \ref{tab:search}, we show that by simply replacing pointwise convolutions in ShuffleNetv2, we are able to outperform state-of-the-art architecture search methods in terms of Top-1 accuracy on ImageNet. We hope that this leads to future work where BFT is included as one of the building blocks in architecture searches, since it provides an extremely low FLOP method for channel fusion.

\begin{figure*}[!t]
\centering
\begin{subfigure}[]{.3\textwidth}
\centering
\includegraphics[width=1.0\linewidth]{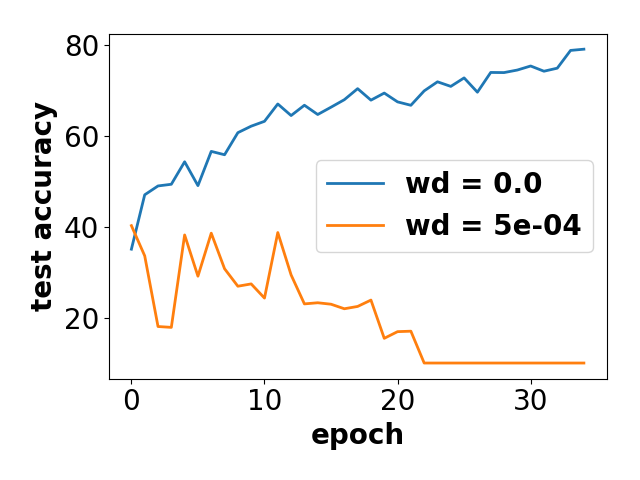}
\captionof{figure}{\textbf{Effect of weight-decay} }
\label{fig:weight_decay}
\end{subfigure}
\begin{subfigure}[]{.3 \textwidth}
\centering
\includegraphics[width=1.0\linewidth]{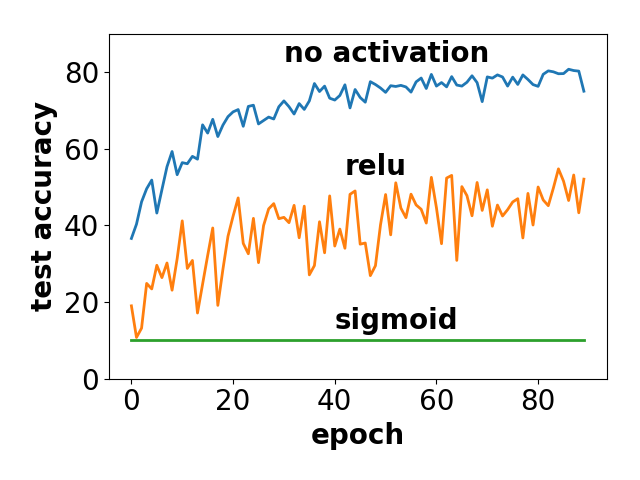}
\captionof{figure}{\textbf{Effect of activations} }
\label{fig:activation}
\end{subfigure}
\begin{subfigure}[]{.3 \textwidth}
\centering
\includegraphics[width=1.0\linewidth]{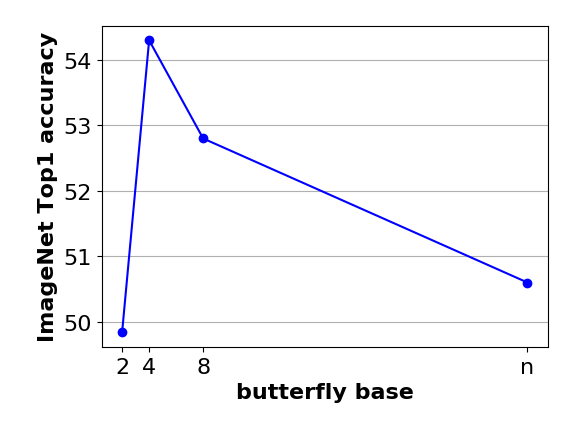}
\captionof{figure}{\textbf{Effect of butterfly base} }
\label{fig:butterfly_base}
\end{subfigure}
\caption{\footnotesize Design choices for BFT: a) In BFT we should not enforce weight decay, because it significantly reduces the effect of input channels on output channels. b) Similarly, we should not apply the common non-linear activation functions. These functions zero out almost half of the values in the intermediate BFLayers, which leads to a catastrophic drop in the information flow from input channels to the output channels. c) Butterfly base determines the structure of BFT. Under $40 M$ FLOP budget base $k=4$ works the best.}
\end{figure*}

\subsubsection{Comparison with Structured Matrices }
To further illustrate the benefits of Butterfly Transforms, we compare them with other structured matrix methods which can be used to reduce the computational complexity of pointwise convolutions. In Table \ref{tab:lowrank} we show that BFT significantly outperforms all these other methods at a similar FLOP range. For comparability, we have extended all the other methods to be used as replacements for pointwise convolutions, if necessary. We then replaced all pointwise convolutions in MobileNetV1 for each of the methods and report Top-1 validation accuracy on ImageNet. Here we summarize these other methods:

\textbf{Circulant block:}
In this block, the matrix that represents the pointwise convolution is a circulant matrix. In a circulant matrix rows are cyclically shifted versions of one another \cite{ding2017c}. The product of this circulant matrix by a column can be efficiently computed in $\mathcal{O}(n \log(n))$ using the Fast Fourier Transform (FFT). 

\textbf{Low-rank matrix:}
In this block, the matrix that represents the pointwise convolution is the product of two $log(n)$ rank matrices ($W = UV^T$). Therefore the pointwise convolution can be performed by two consequent small matrix product and the total complexity is $\mathcal{O}(n \log n)$.

\textbf{Toeplitz Like:}
Toeplitz like matrices have been introduced in \cite{toeplitz_small_footprint}. They have been proven to work well on kernel approximation. We have used displacement rank $r=1$ in our experiments.

\textbf{Fastfood: }
This block has been introduce in \cite{Fastfood41466} and used in Deep Fried ConvNets\cite{Yang2014DeepFC}. In Deep Fried Nets they replace fully connected layers with FastFood. By unifying batch, height and width dimension, we can use a fully connected layer as a pointwise convolution. 

\textbf{BPBP:}
This method uses the butterfly network structure for fast factorization for approximating linear transformation, such as Discrete Fourier Transform (DFT) and the Hadamard transform\cite{dao2019learning}. We extend BPBP to work with pointwise convolutions by using the trick explained in the Fastfood section above, and performed experiments on ImageNet.


\subsection{Ablation Study}
\label{sec:ablation}
Now, we study different elements of our BFT model. As mentioned earlier, residual connections and non-linear activations can be augmented within our BFLayers. Here we show the performance of these elements in isolation on CIFAR-10 dataset using MobileNetv1 as the base network. The only exception is the Butterfly Base experiment which was performed on ImageNet.


\setlength{\intextsep}{5pt}%
\setlength{\columnsep}{7pt}%
\begin{wraptable}{o}{4.5cm}
\begin{tabular}{|l|l|}
\hline
Model               & Accuracy \\ \hline\hline
No residual & 79.2    \\ \hline
Every-other-Layer   & 81.12       \\ \hline
First-to-Last       & \textbf{81.75}     \\ \hline
\end{tabular}
\caption{\textbf{Residual connections}}
\label{tab:residual}
\end{wraptable}

\textbf{Residual connections:}
The graphs that are obtained by replacing BFTransform with pointwise convolutions are very deep. Residual connections generally help when training deep networks. We experimented with three different ways of adding residual connections (1) \emph{First-to-Last}, which connects the input of the first BFLayer to the output of last BFLayer, (2) \emph{Every-other-Layer}, which connects every other BFLayer and (3) \emph{No-residual}, where there is no residual connection. We found the First-to-last is the most effective type of residual connection as shown in Table \ref{tab:residual}.

\textbf{With/Without Non-Linearity:}
As studied by \cite{sandler2018mobilenetv2} adding a non-linearity function like $ReLU$ or $Sigmoid$ to a narrow layer (with few channels) reduces the accuracy because it cuts off half of the values of an internal layer to zero. In BFT, the effect of an input channel $i$ on an output channel $o$, is determined by the multiplication of all the edges on the path between $i$ and $o$. Dropping any value along the path to zero will destroy all the information transferred between the two nodes. Dropping half of the values of each internal layer destroys almost all the information in the entire layer. Because of this, we don't use any activation in the internal Butterfly Layers. Figure \ref{fig:activation} compares the the learning curves of BFT models with and without non-linear activation functions.

\textbf{With/Without Weight-Decay:}  
We found that BFT is very sensitive to the weight decay. This is because in BFT there is only one path from an input channel $i$ to an output channel $o$. The effect of $i$ on $o$ is determined by the multiplication of all the intermediate edges along the path between $i$ and $o$. Pushing all weight values toowards zero, will significantly reduce the effect of the $i$ on $o$. Therefore, weight decay is very destructive in BFT. Figure \ref{fig:weight_decay} illustrates the learning curves with and without using weight decay on BFT.

\textbf{Butterfly base:}
The parameter $k$ in $B^{(n,k)}$ determines the structure of the Butterfly Transform and has a significant impact on the accuracy of the model. The internal structure of the \sys will contain $\log_k(n)$ layers. Because of this, very small values of $k$ lead to deeper internal structures, which can be more difficult to train. Larger values of $k$ are shallower, but have more computation, since each node in layers inside the \sys has an out-degreee of $k$. With large values of $k$, this extra computation comes at the cost of more FLOPs.

We tested the values of $k = 2, 4, 8, n$ on MobileNetV1+BFT with an input resolution of 160x160 which results in $\sim40M$ FLOPs. When $k=n$, this is equivalent to a standard pointwise convolution. For a fair comparison, we made sure to hold FLOPs consistent across all our experiments by varying the number of channels, and tested all models with the same hyper-parameters on ImageNet. Our results in Figure \ref{fig:butterfly_base} show that $k=4$ significantly outperforms all other values of $k$.
Our intuition is that this setting allows the block to be trained easily, due to its shallowness, and that more computation than this is better spent elsewhere, such as in this case increasing the number of channels.
It is a likely possibility that there is a more optimal value for $k$, which varies throughout the model, rather than being fixed. We have also only performed this ablation study on a relatively low FLOP range ($40M$), so it might be the case that larger architectures perform better with a different value of $k$. There is lots of room for future exploration in this design choice.


%

\section{Drawbacks} 
 A weakness of our model is that there is an increase in working memory when using BFT since we must add substantially more channels to maintain the same number of FLOPs as the original network. For example, a MobileNetV1-2.0+BFT has the same number of FLOPS as a MobileNetV1-0.5, which means it will use about four times as much working memory. Please note that the intermediate BFLayers can be computed in-place so they do not increase the amount of working memory needed. Due to using wider channels, GPU training time is also increased. In our implementation, at the forward pass, we calculate $B^{(n,k)}$ from the current weights of the BFLayers, which is a bottleneck in training. Introducing a GPU implementation of butterfly operations would greatly reduce training time. 

\section{Conclusion and Future Work}
\label{sec:conclusion}
In this paper, we demonstrated how a family of efficient transformations referred to as the Butterfly Transforms can replace pointwise convolutions in various neural architectures to reduce the computation while maintaining accuracy. We explored many design decisions for this block including residual connections, non-linearities, weight decay, the power of the \sys, and also introduce a new weight initialization, which allows us to significantly outperform all other structured matrix approaches for efficient channel fusion that we are aware of. We also provided a set of principles for fusion network design, and \sys exhibits all these properties.

As a drop-in replacement for pointwise convolutions in efficient Convolutional Neural Networks, we have shown that our method significantly increases accuracy of models, especially at the low FLOP range, and can enable new capabilities on resource constrained edge devices. It is worth noting that these neural architectures have not at all been optimized for \sys, and we hope that this work will lead to more research towards networks designed specifically with the Butterfly Transform in mind, whether through manual design or architecture search. \sys can also be extended to other domains, such as language and speech, as well as new types of architectures, such as Recurrent Neural Networks and Transformers.


We look forward to future inference implementations of Butterfly structures which will hopefully validate our hypothesis that this block can be implemented extremely efficiently, especially on embedded devices and FPGAs. Finally, one of the major challenges we faced was the large amount of time and GPU memory necessary to train \sys, and we believe there is a lot of room for optimizing training of this block as future work.

\section*{Acknowledgement}
Thanks Aditya Kusupati, Carlo Del Mundo,  Golnoosh Samei, Hessam Bagherinezhad, James Gabriel and Tim Dettmers for their help and valuable comments. 
This work is in part supported by NSF IIS 1652052, IIS 17303166, DARPA N66001-19-2-4031,  67102239  and gifts from  Allen Institute for Artificial Intelligence. 
{\small
\bibliographystyle{ieee_fullname}
\bibliography{BFT}
}

\newpage
\appendix
\section{Experimental details}
Here we explain our experimental setup. For all architectures, we optimize our network by minimizing cross-entropy loss using SGD.
\subsection{MobileNetV1+BFT}
We have used weight decay of $10^{-5}$.  We train for $170$ epochs. We have used a constant learning rate $0.5$ and decay it by $\frac{1}{10}$ at epochs $140, 160$. For details on width multiplier of MobileNet and input resolution on each experiment look at Table \ref{tab:mobilenet_full_table_appendix}.
\subsection{ShuffleNetV2+BFT}
\begin{figure}[t]
    \includegraphics[width= 1.0\linewidth]{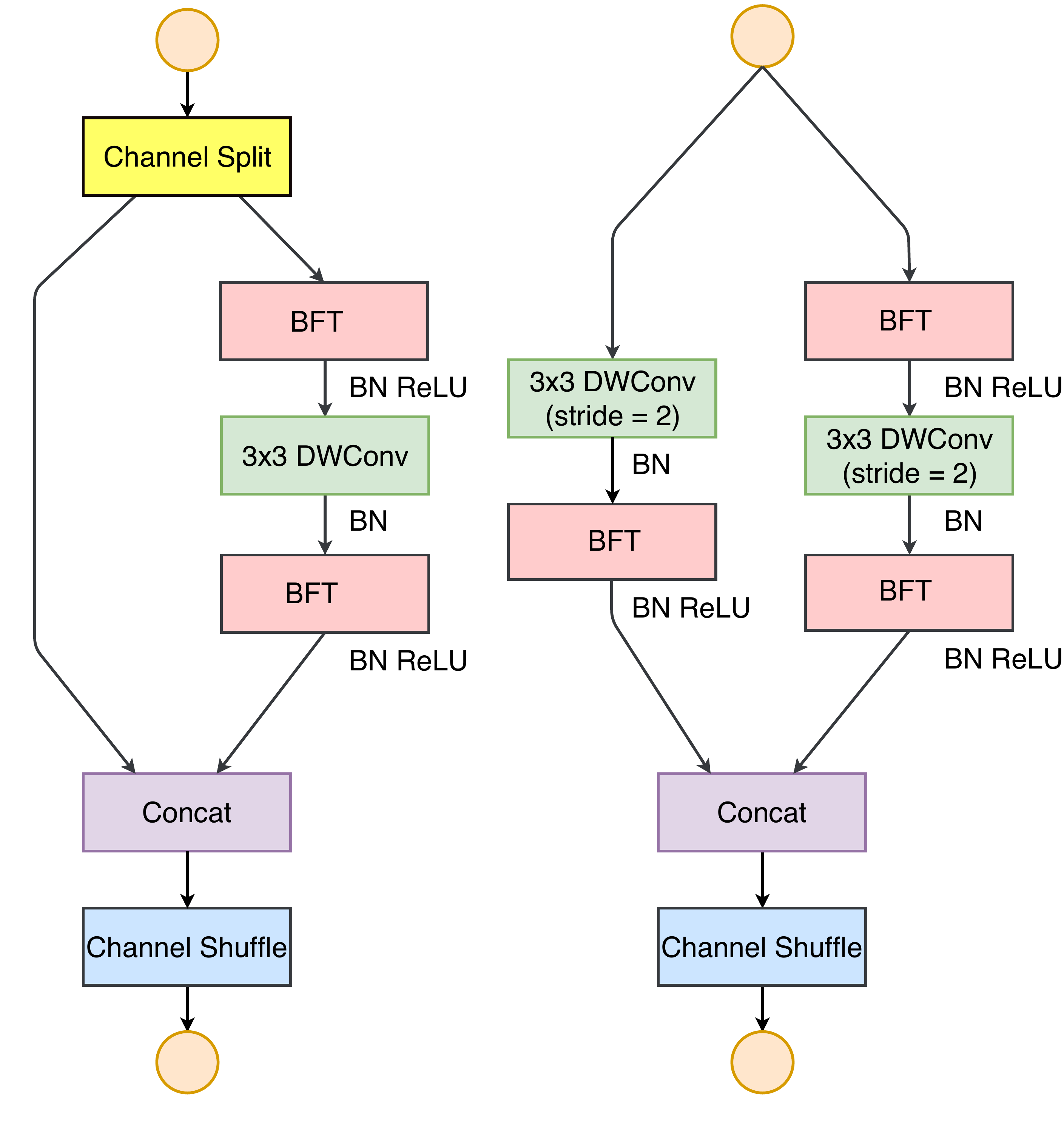}
    \caption {\footnotesize \\ \textbf{ShuffleNetV2+BFT Block}}
    \label{fig:shufflenet_bft_block}
\end{figure}

We have used weight decay of $10^{-5}$.  We train for $300$ epochs. We start with a learning rate of $0.5$ linearly decaying it to $0$. All of the pointwise convolutions are replaced by BFT as shown in Figure \ref{fig:shufflenet_bft_block}, except the first pointwise convolution with input channel size of 24. For comparing under the similar number of FLOPs we have slightly changed ShuffleNet's layer width to create ShuffleNetV2-1.25. This is the structure which is used for shuffleNetV2-1.25:
\begin{table}[h]
\begin{small}
\begin{tabular}{|l|l|l|l|l|l|}
\hline
Layer                                                    & output size                                             & Kernel                                            & Stride                                        & Repeat                                        & Width \\ \hline
Image                                                    & 224$\times$224                                                 &                                                   &                                               &                                               & 3    \\ \hline
\begin{tabular}[c]{@{}l@{}}Conv1\\ Max pool\end{tabular} & \begin{tabular}[c]{@{}l@{}}112$\times$112\\ 56$\times$56\end{tabular} & \begin{tabular}[c]{@{}l@{}}3$\times$3\\ 3$\times$3\end{tabular} & \begin{tabular}[c]{@{}l@{}}2\\ 2\end{tabular} & 1                                             & 24   \\ \hline
Stage 2                                                  & \begin{tabular}[c]{@{}l@{}}28$\times$28\\ 28$\times$28\end{tabular}   &                                                   & \begin{tabular}[c]{@{}l@{}}2\\ 1\end{tabular} & \begin{tabular}[c]{@{}l@{}}1\\ 3\end{tabular} & 128  \\ \hline
Stage 3                                                  & \begin{tabular}[c]{@{}l@{}}14$\times$14\\ 14$\times$14\end{tabular}   &                                                   & \begin{tabular}[c]{@{}l@{}}2\\ 1\end{tabular} & \begin{tabular}[c]{@{}l@{}}1\\ 7\end{tabular} & 256  \\ \hline
Stage 4                                                  & \begin{tabular}[c]{@{}l@{}}7$\times$7\\ 7$\times$7\end{tabular}       &                                                   & \begin{tabular}[c]{@{}l@{}}2\\ 1\end{tabular} & \begin{tabular}[c]{@{}l@{}}1\\ 3\end{tabular} & 1024 \\ \hline
Conv 5                                                   & 7$\times$7                                                     & BFT                                               & 1                                             & 1                                             & 1024 \\ \hline
Global Pool                                              & 1$\times$1                                                     & 7$\times$7                                               &                                               &                                               &      \\ \hline
FC                                                       &                                                         &                                                   &                                               &                                               & 1000 \\ \hline
FLOPS                                                    &                                                         &                                                   &                                               &                                               & 41 \\ \hline
\end{tabular}
\end{small}
\end{table}

For details on input resolution on each experiment look at Table \ref{tab:shufflenet_full_table}.

\subsection{MobileNetV3+BFT}
We have used weight decay of $10^{-5}$.  We train for $200$ epochs. We start with a warm-up for the first $5$ epochs, starting from a learning rate $0.1$ and linearly increasing it to $0.5$. Then we decay learning rate from $0.5$ to $0.0$ using a cosine scheme in the remaining $195$ epochs.
For details on width multiplier and input resolution on each experiment look at Table \ref{tab:mobilenetv3_full_table}.

\begin{table*}[!t]
        \centering
        \begin{tabular}{|l|l|l|l||l|l|l|l||l|}
            \hline
            \multicolumn{4}{|c||}{MobileNet} & \multicolumn{4}{|c||}{MobileNet\textbf{+BFT}} & \multirow{2}{*}{gain} \\  \cline{1-8}
            width & resolution & flops & Accuracy & width & resolution & flops & Accuracy &  \\ \hline \hline
            0.25 & 128 & 14 M & 41.50 & 1.0 & 96 & 14 M & 46.58 & \textbf{5.08}\\ \hline
            0.25 & 160 & 21 M & 45.50 & 1.0 & 128 & 23 M & 52.26 & \textbf{6.76} \\ \hline
            0.25 & \begin{tabular}[c]{@{}l@{}} 192 \\ 224 \\ \end{tabular}& \begin{tabular}[c]{@{}l@{}}34 M \\ 41 M\\ \end{tabular} & \begin{tabular}[c]{@{}l@{}}47.70 \\ 50.60 \\ \end{tabular}
            & 1.0 & 160 & 35 M & 54.30 & \begin{tabular}[c]{@{}l@{}} \textbf{6.60} \\ \textbf{3.70} \end{tabular} \\ \hline
            0.50 & 128 & 49 M & 56.30 & \begin{tabular}[c]{@{}l@{}} 1.0 \\ 2.0\\ \end{tabular} & \begin{tabular}[c]{@{}l@{}}192 \\ 128\\ \end{tabular} & \begin{tabular}[c]{@{}l@{}}51 M \\ 52 M\\ \end{tabular} & \begin{tabular}[c]{@{}l@{}}57.56 \\ 58.35 \\ \end{tabular} & \begin{tabular}[c]{@{}l@{}} \textbf{1.26} \\ \textbf{2.05} \end{tabular} \\ \hline
            0.50 & 192 & 110 M & 61.70 & 2.0 & 192 & 112 M & 63.03 & \textbf{1.33} \\ \hline
            0.50 & 224 & 150 M & 63.30 & 2.0 & 224 & 150 M & 64.32 & \textbf{1.02} \\ \hline
        \end{tabular}
        \caption{\footnotesize Comparision between Mobilenet and Mobilenet+BFT. For comparision under similar number of FLOPs we have used wider channels in MobileNet+BFT.}
        \label{tab:mobilenet_full_table_appendix}
\end{table*}

\begin{table*}[!t]
        \centering
        \begin{tabular}{|l|l|l|l||l|l|l|l||l|}
            \hline
            \multicolumn{4}{|c||}{ShuffleNetV2} & \multicolumn{4}{|c||}{ShuffleNetV2\textbf{+BFT}} & \multirow{2}{*}{gain} \\  \cline{1-8}
            width & resolution & flops & Accuracy & width & resolution & flops & Accuracy &  \\ \hline \hline
            0.50 & 128 & 14 M & 50.86* & 1.25 & 128 & 14 M & 55.26 & \textbf{4.4}\\ \hline
            0.50 & 160 & 21 M & 55.21* & 1.25 & 160 & 21 M & 57.83 & \textbf{2.62} \\ \hline
            0.50 & 224 & 41 M & \begin{tabular}[c]{@{}l@{}}59.70*\\ 60.30\end{tabular} & 1.25 & 224 & 41 M & 61.33 & \begin{tabular}[c]{@{}l@{}}\textbf{1.63}\\\textbf{1.03}\end{tabular} \\ \hline
        \end{tabular}
        \caption{\footnotesize Comparision between ShuffleNetV2 and ShuffleNetV2+BFT. For comparision under similar number of FLOPs we have used wider channels in ShuffleNetV2+BFT.}
        \label{tab:shufflenet_full_table}
\end{table*}

\begin{table*}[!t]
        \centering
        \begin{tabular}{|l|l|l|l||l|l|l|l||l|}
            \hline
            \multicolumn{4}{|c||}{MobileNetV3} & \multicolumn{4}{|c||}{MobileNetV3\textbf{+BFT}} & \multirow{2}{*}{gain} \\  \cline{1-8}
            width & resolution & flops & Accuracy & width & resolution & flops & Accuracy &  \\ \hline \hline
            Small-0.35 & 224 & 13 M & 49.8 & Small-0.5 & 224 & 15 M & 55.21 & \textbf{5.41} \\ \hline
        \end{tabular}
        \caption{\footnotesize Comparision between MobileNetV3 and MobileNetV3+BFT. For comparision under similar number of FLOPs we have used wider channels in MobileNetV3+BFT.}
        \label{tab:mobilenetv3_full_table}
\end{table*}

.
\end{document}